\documentclass[sigplan, 10pt]{acmart}
\AtBeginDocument{%
  \providecommand\BibTeX{{%
    \normalfont B\kern-0.5em{\scshape i\kern-0.25em b}\kern-0.8em\TeX}}}

\acmYear{2023}\copyrightyear{2023}
\setcopyright{rightsretained}
\acmConference[EuroMLSys '23]{3rd Workshop on Machine Learning and Systems}{May 8, 2023}{Rome, Italy}
\acmBooktitle{3rd Workshop on Machine Learning and Systems (EuroMLSys '23), May 8, 2023, Rome, Italy}
\acmPrice{}
\acmDOI{10.1145/3578356.3592577}
\acmISBN{979-8-4007-0084-2/23/05}

\usepackage{caption}
\usepackage{subcaption}

\begin{document}

%%
%% The "title" command has an optional parameter,
%% allowing the author to define a "short title" to be used in page headers.
\title{MCTS-GEB: Monte Carlo Tree Search is a Good E-graph Builder}

\author{Guoliang He}
\email{gh512@cam.ac.uk}
\affiliation{
  \institution{University of Cambridge}
  %\streetaddress{1 Th{\o}rv{\"a}ld Circle}
  %\city{Hekla}
  \country{United Kingdom}
  }

\author{Zak Singh}
\email{zs391@cantab.ac.uk}
\affiliation{
  \institution{University of Cambridge}
  %\streetaddress{1 Th{\o}rv{\"a}ld Circle}
  %\city{Hekla}
  \country{United Kingdom}
  }

\author{Eiko Yoneki}
\email{eiko.yoneki@cl.cam.ac.uk}
\affiliation{
  \institution{University of Cambridge}
  %\streetaddress{1 Th{\o}rv{\"a}ld Circle}
  %\city{Hekla}
  \country{United Kingdom}
  }

\begin{abstract}

Rewrite systems \cite{herbie, spores, tensat} have been widely employing equality saturation \cite{es}, which is an optimisation methodology that uses a saturated e-graph to represent all possible sequences of rewrite simultaneously, and then extracts the optimal one. As such, optimal results can be achieved by avoiding the phase-ordering problem. However, we observe that when the e-graph is not saturated, it cannot represent all possible rewrite opportunities and therefore the phase-ordering problem is re-introduced during the construction phase of the e-graph. To address this problem, we propose \textbf{MCTS-GEB}, a domain-general rewrite system that applies reinforcement learning (RL) to e-graph construction. At its core, MCTS-GEB uses a Monte Carlo Tree Search (MCTS) \cite{mcts} to efficiently plan for the optimal e-graph construction, and therefore it can effectively eliminate the phase-ordering problem at the construction phase and achieve better performance within a reasonable time. Evaluation in two different domains shows MCTS-GEB can outperform the state-of-the-art rewrite systems by up to $49$x, while the optimisation can generally take less than an hour, indicating MCTS-GEB is a promising building block for the future generation of rewrite systems.

\end{abstract}

  

%%
%% Keywords. The author(s) should pick words that accurately describe
%% the work being presented. Separate the keywords with commas.

%\received{20 February 2007}
%\received[revised]{12 March 2009}
%\received[accepted]{5 June 2009}

%%
%% This command processes the author and affiliation and title
%% information and builds the first part of the formatted document.
\maketitle

\section{Introduction}
\label{seq. intro}

A rewrite system typically applies a sequence of rewrites to transform the intermediate representation (IR) of the target program, so that the target program is transformed to a more optimised form. However, applying rewrites may be destructive, meaning a bad rewrite can hide opportunities for subsequent program transformation because it replaces some part of the IR with a new one, as specified by the rewrite rule. Furthermore, rewrites can be dependent on one another. Therefore finding the optimal sequence of rewrites is difficult and it is collectively known as the "phase-ordering" problem in compiler optimisation. 

Equality saturation \cite{es} is proposed to address the phase-ordering problem by utilising a data structure, the equivalence graph (e-graph). In contrast to applying rewrites sequentially, equality saturation attempts to apply rewrite all at once, and represent those rewrite opportunities in an e-graph. Thus, the phase-ordering problem can be avoided since all rewrites are simultaneously represented by the e-graph. Such an e-graph is called a \textit{saturated} e-graph, meaning it can no longer grow because it has already represented all possible rewrite opportunities.

However, we notice that e-graphs can rarely be \textit{saturated} for many real-world scenarios. This is because e-graphs can grow unbounded during the construction phase, and eventually may be too large to be stored in memory. In addition, finding the optimal rewrite sequence from a very large e-graph can be time-consuming, since the extractor has to traverse the e-graph and build up the optimised IR. As a result, people often set an upper-bound node limit to the e-graph. For example, EGG \cite{egg} is an efficient domain-general equality saturation library and it upper-bounds the number of e-nodes by 10,000 by default.

As a result, the phase-ordering problem is re-introduced during the construction phase because the e-graph is not built to saturation. To overcome this problem, we propose \textbf{MCTS-GEB}, a domain-general rewrite system that leverages MCTS for efficient and optimised e-graph construction. MCTS is a specific reinforcement learning algorithm, which is a general optimising framework for sequential decision-making problems. Therefore, MCTS-GEB can plan for the optimal sequence of rewrite rules to build the e-graph, instead of just applying all rewrite rules until the e-node limit is met.

Applying reinforcement learning (RL) to rewrite systems has several challenges. First, the e-graph may grow unrestricted and contain up to tens of thousands of nodes, which is hard to embed and make decisions efficiently in a single machine. Second, a typical rewrite system should be able to finish optimisation within a reasonable time, while training RL agents can take hours or even days from scratch, making common RL agents an infeasible solution. Third, the feedback signal for RL is important as RL is meant to maximise the expected cumulative reward. However, it can only be generated by traversing the e-graph and extracting the optimised IR. Providing a feedback signal for each step may bring significant overhead to the rewrite system, so the feedback signal may be sparse. 

We choose MCTS and modify it to address the above challenges. MCTS is a planning-based method meaning it does not need to embed the e-graph, so it can efficiently scale with e-graph sizes. Moreover, MCTS's search-based planning can be efficiently accelerated by hardware parallelism. Finally, MCTS is proven to be successful to deal with sparse rewards, as shown by the chess-playing agent AlphaGo \cite{alphago}. As a result, we decide to adapt MCTS to equality saturation. MCTS-GEB is available as open source\footnote{\url{https://github.com/ucamrl/eqs.git}}.

In summary, this paper makes the following contributions:

\begin{itemize}
    \item We introduce MCTS-GEB, a domain-general rewrite system that leverages MCTS for efficient and optimised equality saturation.
    \item Our evaluation covers two different domains and shows MCTS-GEB can outperform the state-of-the-art by up to $49$x, while the optimisation time takes less than an hour.
\end{itemize}

\section{Background and motivation}

\subsection{Equality saturation}

In this section, we introduce the core idea of equality saturation. We use the same terminologies as in EGG \cite{egg} for consistency. Typically, rewrite systems take as input an initial IR of the target program and a set of rewrite rules as specified by programmers, where each rewrite rule is denoted as $l \rightarrow r$. Both $l$ and $r$ are patterns and the rewrite rule indicates their equivalency. For example, the commutative property of addition may be represented by the rewrite rule $(a+b) \rightarrow (b+a)$, where both $a$ and $b$ may match any variable in the target IR. To perform rewrites, the system iteratively applies rewrite rules to the target IR. For each rewrite rule $l \rightarrow r$, there may be a sequence of matches to the target IR, where each match $\sigma$ maps the abstract pattern $l$ to a specific variable. Then applying $\sigma[r]$ will transform the matched variable according to the rewrite rule, and therefore the target IR is rewritten. 

\begin{figure}
    \centering
    \includegraphics[width=\linewidth]{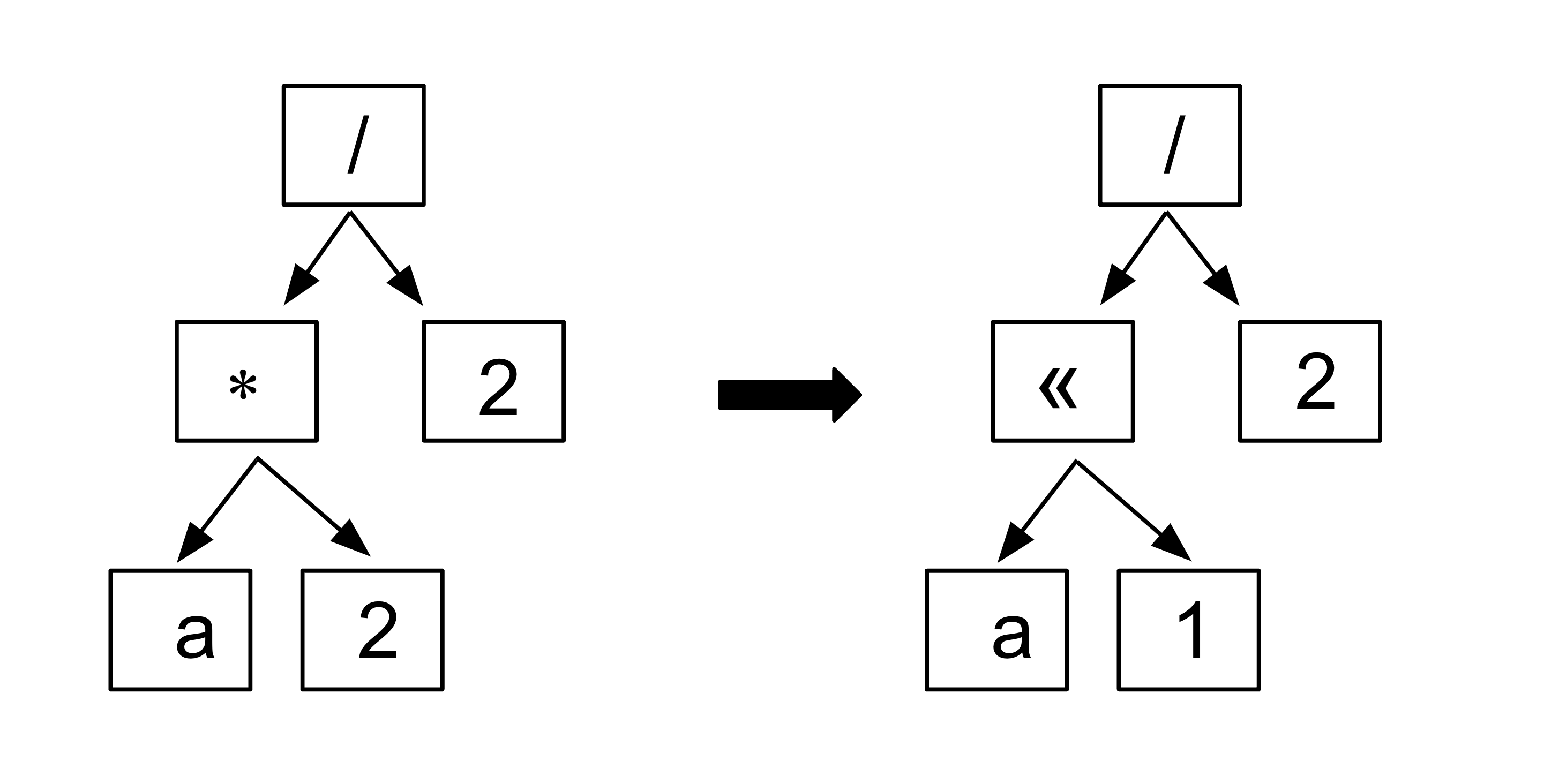}
    %\vspace*{-20mm}
    \caption{A term rewriting example. The expression on the left-hand side is the initial IR, and applying $(x\times 2) \rightarrow (x<< 1)$ gives the transformed result on the right-hand side.}
    \label{fig: A term rewriting example}
\end{figure}

A classic example that illustrates this process is simplifying the expression of $(a\times2 )/2$. As shown by Figure \ref{fig: A term rewriting example}, after applying a rewrite rule $(x\times 2) \rightarrow (x<< 1)$, the rewrite system will find a match $\sigma = \{ x \longmapsto a\}$, and it leads to the result $(a<<1)/2$.

%\vspace*{-20mm}
%\begin{figure}[H]

%\begin{figure}
%     \centering
%     \begin{subfigure}[b]{0.4\linewidth}
%         \centering
%         \includegraphics[width=\linewidth]{images/term-rewrite-example.png}
%         \caption{$y=x$}
%         \label{fig:y equals x}
%     \end{subfigure}
%     \hfill
%     \begin{subfigure}[b]{0.4\linewidth}
%         \centering
%         \includegraphics[width=\linewidth]{images/term-rewrite-example.png}
%         \caption{$y=3\sin x$}
%         \label{fig:three sin x}
%     \end{subfigure}
%\end{figure}

\begin{figure}[ht]
    \centering
    \includegraphics[width=\linewidth]{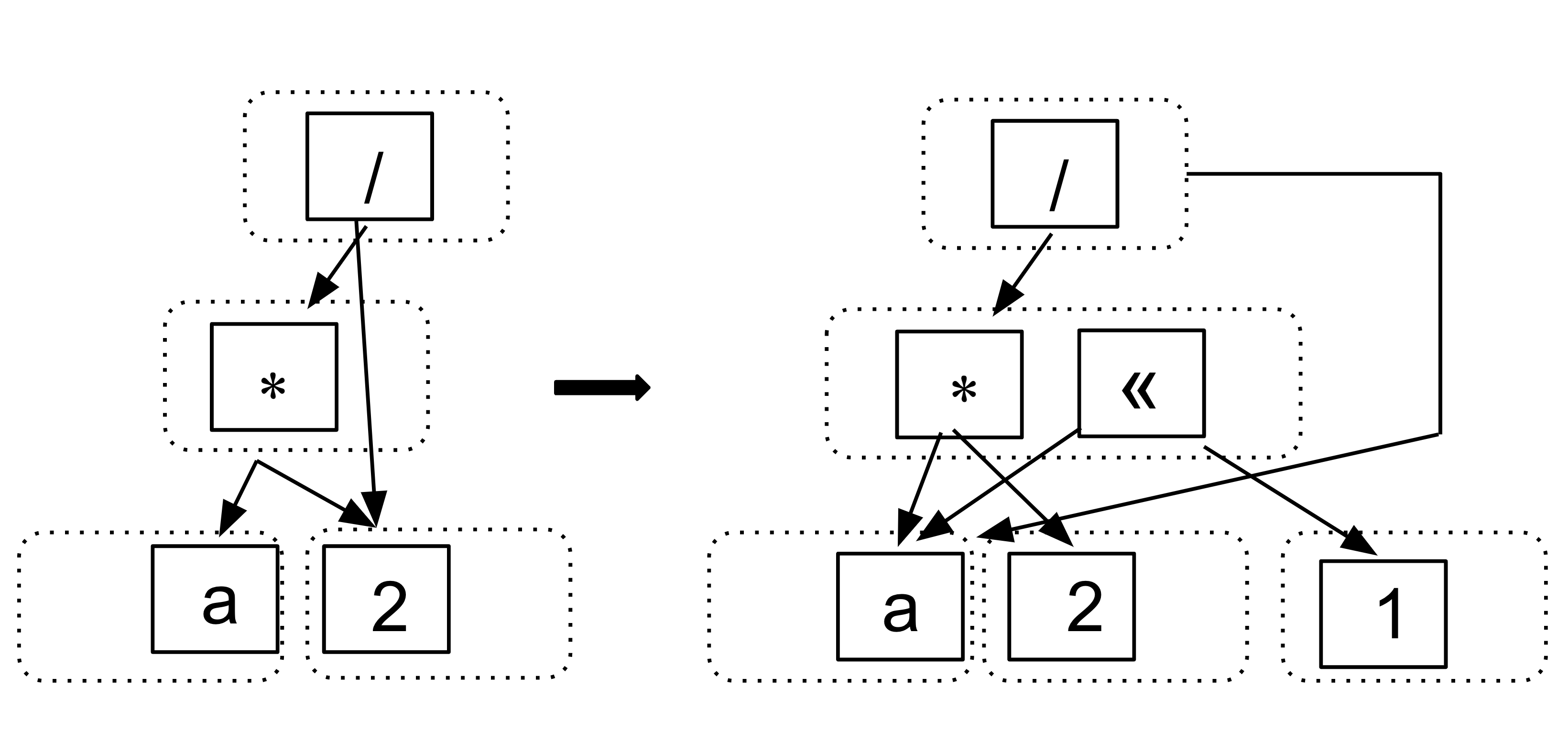}
    %\vspace*{-22mm}
    \caption{An e-graph example. The expression on the left-hand side is the initial IR represented by an e-graph, and applying $(x\times 2) \rightarrow (x<< 1)$ gives the transformed result on the right-hand side, but unlike Figure \ref{fig: A term rewriting example}, we can see that the transformed IR does not lose information.}
    \label{fig: An e-graph example}
\end{figure}

The example in Figure \ref{fig: A term rewriting example} also shows the limitation of rewriting systems. It is easy to see that the optimal result should be just $a$, but applying a "wrong" rewrite rule may hide the opportunity to transform the IR optimally. Hence the idea of equality saturation is to use an e-graph to simultaneously represent all possible rewrite opportunities. An e-graph representation of the same example is visualised by Figure \ref{fig: An e-graph example}. 

The e-graph representation has two kinds of nodes, e-class and e-node. In Figure \ref{fig: An e-graph example}, the square nodes are e-nodes, and the dotted nodes are e-classes. An e-node represents a symbol in the target IR, whereas an e-class may contain a group of e-nodes, indicating those e-nodes are \textit{equivalent}. In this case, we can observe that applying rewrites to the target IR is not destructive, meaning the old representation (the multiplication $*$) still exists after applying the rewrite rule. Therefore the rewrite system does not "forget" information, and can avoid the phase-ordering problem. The effectiveness of equality saturation has been proven by rewrite systems across different domains, such as floating point number accuracy improvement \cite{herbie}, relational algebraic rewrite \cite{spores}, and tensor graph super-optimisation \cite{tensat}.

%\vspace*{-20mm}

\subsubsection{Extraction} Since different rewrite sequences co-exist in the e-graph, an extractor is needed to extract the optimal sequence from the e-graph. The extractor traverses the e-graph starting from a root node and assigns each node a cost based on a customisable cost function. The extractor can use a greedy strategy or an integer linear programming (ILP) \cite{ilp} strategy to compute the desired path and perform an extraction.  

\subsubsection{Limitation}

While Figure \ref{fig: An e-graph example} shows e-graphs can simultaneously represent all rewrite opportunities, it also means each rewrite only adds nodes to the e-graph. Thus, an e-graph may grow unbounded during the construction phase, and people have to upper-bound the node limit of an e-graph to prevent it runs out of memory. For example, as soon as the e-graph has more than $10,000$ e-nodes, EGG \cite{egg} stops the construction. This indicates \textit{how the e-graph is built matters}. Therefore, the phase-ordering problem is introduced during the construction phase. 

\subsection{Reinforcement learning}
\label{seq. rl}

Reinforcement learning (RL) \cite{rl2, bellman1957} is a framework for optimising sequential decision-making problems. In this framework, an agent is designed to interact with the environment over several iterations in an episode. The goal of the agent is to learn to maximise the expected cumulative reward collected from the episode. A Markov decision process (MDP) \cite{mdp1, mdp2} is often introduced to specify the RL elements, as shown in the following:

\begin{itemize}
  \item $\mathcal{S}$, is the state space, which consists of the set of valid states.
  \item $\mathcal{A}$, is the action space, which consists of the set of valid actions.
  \item $\mathcal{P}_a$, is the transition probability function that takes as input an action $a$ and a state $s_t$ and makes a transition to state $s'_{t+1}$.
  \item $\mathcal{R}_a$, is the reward function, it returns the reward from the environment after taking an action $a$ between state $s_t$ and $s'_{t+1}$.
\end{itemize}

Among various RL algorithms, we employ MCTS \cite{mcts, mcts2} in this work, because it does not include an expensive training process, which could take hours. Instead, before taking action at each iteration, MCTS builds a search tree to reason about what is the optimal action for maximising the long-term reward. As a result, MCTS achieves a good balance between decision-making and optimisation time, so we consider it a good fit for our purpose.

\subsubsection{Monte Carlo Tree Search}

MCTS is a model-based RL algorithm that performs look-ahead planning via its environment transition model. At each planning stage, it performs the following four steps iteratively.

\begin{enumerate}
  \item selection: it traverses down from the tree node $R$ until a leaf node $L$ is reached. 
  \item expansion: a new tree node $C$ is expanded from the leaf node $L$.
  \item simulation: it performs a rollout from node $C$, and records the reward.
  \item backup: it uses the reward to update the value estimate from node $C$ up to $R$.
\end{enumerate}

% can have more details about UCT selection

The planning stage ends when the search budget is exhausted, which is a configurable hyperparameter in our system. Upon finishing planning, the agent selects the next action which corresponds to the maximal value child node of root node $R$.

While MCTS is a well-established algorithm, its wide application is hindered by the dependency on an environment transition model for planning. However, this is not a problem in our case, because we can control the construction phase completely by interfacing with EGG \cite{egg}. Therefore, the environment transition model is perfectly known and MCTS can be applied readily. 

\subsubsection{Parallel Monte Carlo Tree Search}

MCTS is a single-thread algorithm, meaning it executes the four planning steps sequentially. To improve computational efficiency, parallel MCTS \cite{pmcts1, pmcts2} algorithms are proposed to build the search tree in parallel. Specifically, multiple workers can be dispatched to simultaneously perform expansion and simulation. Inevitably, this may violate the selection policy because the latest global statistics for selection are only available after the backup stage is done. To mitigate this problem, WU-UCT \cite{wu-uct} is a recent work to introduce new statistics for selection policy. Over a range of benchmarks, it achieves similar results as the single-thread MCTS while having a much lower search time. As such, MCTS-GEB adopts the implementation of WU-UCT.

\section{MCTS-GEB}

In this section, we introduce MCTS-GEB, a domain-general rewrite system that leverages MCTS for efficient and optimised equality saturation. 

%\vspace*{-3mm}
\begin{figure}[ht]
    \centering
    \includegraphics[width=\linewidth]{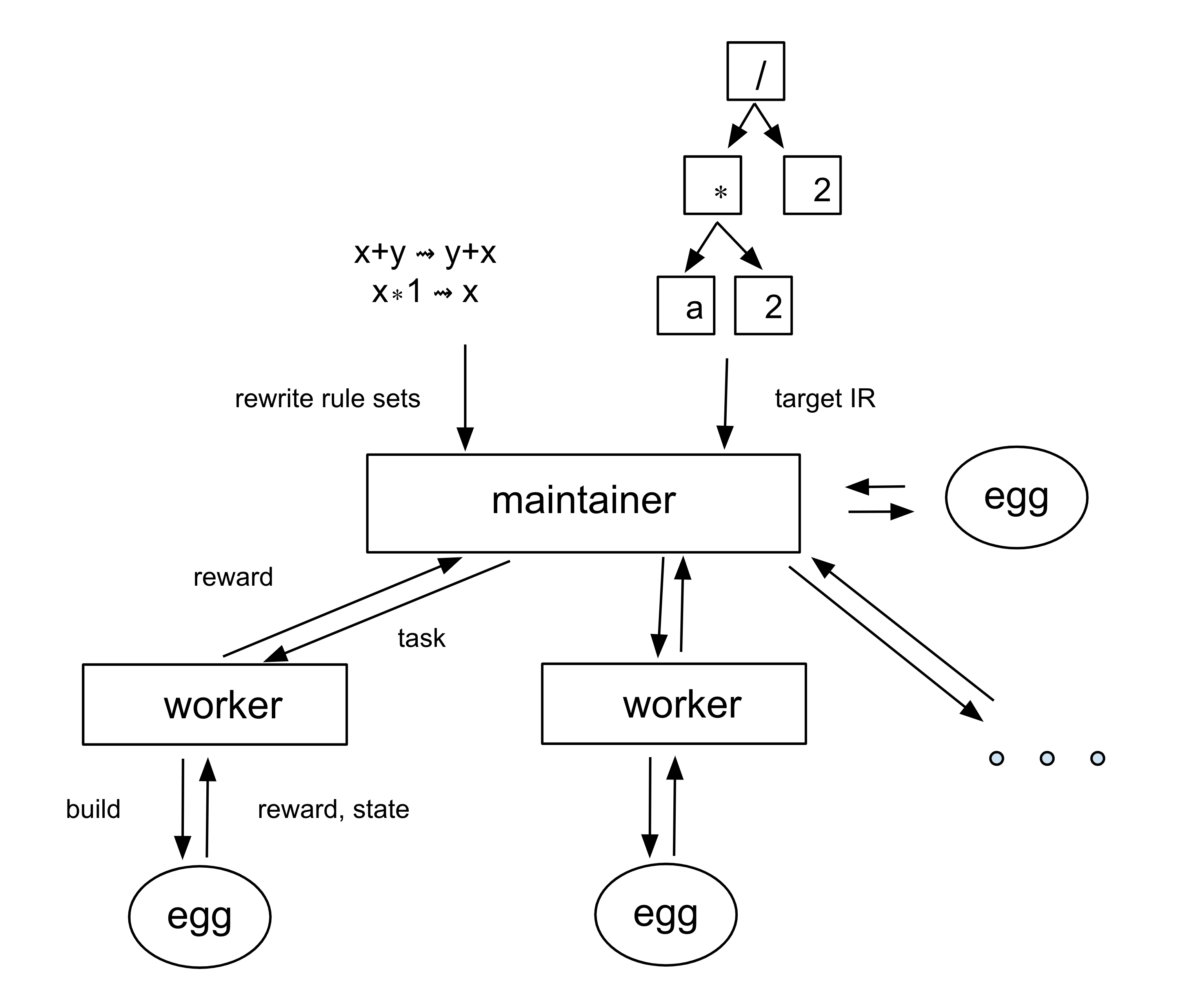}
    %\vspace*{-20mm}
    \caption{The overview of MCTS-GEB, where square nodes denote processes and circle nodes are the EGG library. Their interactions are shown by arrows.}
    \label{fig: The overview of MCTS-GEB}
\end{figure}
%\vspace*{-3mm}

Figure \ref{fig: The overview of MCTS-GEB} shows the overview of MCTS-GEB. MCTS-GEB takes as input an initial IR of the target program and a set of rewrite rules as specified by the programmer, and outputs the equivalent and optimised IR. MCTS-GEB has a maintainer to centrally maintain the MCTS search tree, and the maintainer dispatches expansion as well as simulation tasks to workers. Upon receiving the target IR and the rewrite rule set, the maintainer initialises its e-graph and launches multiple workers, each of which also initialises a local e-graph through the EGG library \cite{egg}. Then the construction phase begins and is performed iteratively. 

At each iteration of construction, the maintainer's local e-graph corresponds to the root node $R$, and it performs a selection step and dispatch expansion as well as simulation tasks to workers. After completion, a new tree node is added to the MCTS and its value estimate is attached. In other words, the maintainer only performs selection and backup, while the expansion and simulation tasks are enqueued and dispatched to workers, so the four MCTS steps can be performed asynchronously. When the search budget is exhausted, the maintainer selects the maximal value child node of the root node $R$, and the corresponding action is made to update its e-graph. Then the MCTS search tree is discarded, and another construction step starts again from the updated e-graph. In the end, when the maintainer's e-graph hits the node limit, the construction phase ends and an extractor is used to extract the optimal IR from the maintainer's e-graph.

\subsection{RL formulation}

We formulate the e-graph construction phase following the MDP specification as demonstrated in section \ref*{seq. rl}.

\subsubsection*{E-graph environment} 

MCTS-GEB uses the EGG library to manipulate e-graphs. EGG has a domain-general e-graph implementation, such that users may define their customised languages and build an e-graph to represent the IR of the language. The e-graph construction phase is then encapsulated as an OpenAI gym \cite{brockman2016openai} environment, which provides a standardised interface for a wide range of algorithms. Among all RL algorithms, we specifically choose MCTS for its efficiency.

%For example, EGG has a runner module that exposes a \texttt{with\_expr()} API which initialises an e-graph with the target IR. To apply rewrite rules, it has a \texttt{run()} function, which takes as input the rewrite rule set and builds the initialised e-graph until some stop conditions are met. Therefore, we can control the e-graph construction by supplying one rewrite rule with one iteration limit at a time. The e-graph construction phase is then encapsulated as an OpenAI gym \cite{brockman2016openai} environment, which provides a standardised interface for a wide range of algorithms. Among all RL algorithms, we specifically choose MCTS for its efficiency.

\subsubsection*{State-Action space} 

At each iteration of the construction, MCTS-GEB may choose one of the rewrite rules from the rewrite rule set to build the e-graph. Therefore, we index all rewrite rules and map them to discrete action IDs. This indicates the action space is fixed and is completely determined by the rewrite rule set.

The state space consists of all possible valid e-graphs during the construction phase. However, this introduces a large memory footprint because each valid e-graph corresponds to an MCTS tree node, and each expansion step will add a tree node to the MCTS search tree. Thus, different e-graphs are frequently written to memory during the planning stage. Besides, when the maintainer dispatches an expansion or simulation task to a worker, the worker needs to know the beginning e-graph of the task. This means e-graphs are recurrently transferred by inter-process connection. Substantial overhead is therefore introduced by repeatedly serialising and de-serialising e-graphs.

We adopt an approach that does not represent the e-graphs explicitly in memory. Instead, we represent an e-graph by the action sequence that it is built from. This is because the building process is deterministic once the rewrite rules and the initial expression are given. As a result, we can always reproduce an e-graph given its action sequence. Specifically, each tree node is assigned to a unique index, which is mapped to an action sequence. The maintainer always dispatches a task with a corresponding action sequence. Upon receiving the task, workers will replay the action sequence and obtain the starting e-graph to perform its task.

\subsubsection*{Reward function}

The reward function will guide MCTS towards higher reward regions, and therefore it should be carefully designed to encourage better e-graph construction. We define that an e-graph is better than another if the extractor can extract a better IR from the e-graph.

Extraction can be formulated as an e-graph transversal problem, which starts from the root e-class and selects exactly one e-node from its children. Then chosen the e-node may point to other e-classes, and the selection step is conducted for each of those e-classes again until no more e-class is available for selection. In other words, the extraction phase is similar to a recursive descent of the e-graph. 

To make a selection among the children of an e-class, a cost for each e-node is needed. This is achieved via a cost function, where EGG exposes it as a call-back function for each symbol of the target IR. The default cost function is simply $1$ for each symbol, so the extractor will treat each symbol equally and attempt to find the IR with the lowest length.

The selection policy can employ two different strategies, greedy selection or integer linear programming (ILP). In greedy selection, the extractor will always select the lowest cost when choosing among e-nodes. For ILP, an ILP solver is used to extract the optimal IR. The two different strategies introduce a trade-off between extraction speed and extraction performance.

With that, MCTS-GEB allows users to register a call-back function to compute the reward. It takes as input one initial IR as well as one extracted IR from the current e-graph and outputs the reward. For example, our default cost function is:

\begin{equation}
  R = \max(\text{init\_cost} - \text{current\_cost}, 0)
\end{equation}

The reward function simply computes the difference between the initial IR and the extracted IR, and it is lower-bounded by $0$. Note that by default, this reward function is only called at the end of the simulation step, meaning the feedback signal is sparse. 

\section{Implementation}

MCTS-GEB is built on top of the EGG \cite{egg} library, which is used for e-graph construction and extraction. We also adopt the implementation from WU-UCT \cite{wu-uct} as the parallel MCTS algorithms. In addition, we implement several optimisations for efficient MCTS planning.

\subsubsection*{Action space pruning}

Pruning can be performed because applying some rewrite rules to the e-graph may not introduce any e-node, meaning those rewrite rules cannot match any pattern. Thus, we should filter out those rewrite rules as viable options in the planning phase to avoid redundant computation. When the maintainer dispatches expansion tasks to workers, the tasks are associated with a flag indicating whether or not it is the children of the root node $R$. If the flag is true and the expansion does not add any e-node to the e-graph, the worker returns a saturated flag, and the maintainer will exclude the corresponding action at the end of the planning phase.

\subsubsection*{Straggler respawn}

Stragglers may exist because workers execute tasks concurrently. At the end of the planning phase, we observe that substantial time is wasted by straggler processes to finish their tasks and hence slow down the entire planning phase. As a result, after dispatching the final task, the maintainer simply timeouts in a few seconds and respawns stragglers. This will inevitably cause some tree node loss, however, the loss is small compared to the overall MCTS search tree ($512$ nodes by default), and it saves a significant amount of planning time.

\subsubsection*{Simulation early stopping}

While the simulation tasks performed by workers are meant to roll out to the end of the episode and get the final rewards, we observe that those long-tail simulation steps contribute negligibly to the value estimates, because their rewards are discounted to almost zero. At the same time, they may bring significant overhead to the initial planning phase, which can have up to hundreds of simulation steps ahead. As a result, we cap the maximum simulation step with a hyperparameter \textit{max\_sim\_step} and expose it to users. In our evaluation, we use the default value $20$, and keep it fixed across our benchmarks.

\section{Evaluation}

In this section, we aim to evaluate MCTS-GEB to answer the following questions:

\begin{itemize}
  \item Can MCTS-GEB obtain better expressions compared to normal equality saturation-based rewrite systems?
  \item How much optimisation time is needed for MCTS-GEB?
\end{itemize}

\subsubsection*{benchmarks setup} The EGG repository\footnote{\url{https://github.com/egraphs-good/egg}} provides benchmark suites across a wide range of domains, and each domain has an associated EGG-based system. For example, the \textit{Math} suite implements a rewrite system to simplify the mathematical expression and the \textit{Prop} suite tests how propositional logic can be simplified. Thus, we take the rewrite rule sets from the two domains and randomly generate to-be-optimised expressions. To generate expressions, we employ a depth-first-search fashion to recursively descend expression trees. For each node, we randomly pick a symbol from the domain-specific language and then move on to the child of the node if necessary, until the tree reaches a maximum depth.

\subsubsection*{Experiment setup} We evaluate MCTS-GEB in a server running Ubuntu Linux 20.04 with a 24-core Intel $2.00$GHz E5-2620, and has 256GB RAM. We use the default maximum e-node limit of $10,000$ to prevent e-graphs from growing unbounded. For MCTS-GEB, we use $22$ workers to perform simulation and just one worker to perform expansion. All hyperparameters are fixed across benchmarks.

\subsection{End-to-end expression simplification}

\begin{figure}[ht]
  \centering
  \includegraphics[width=\linewidth]{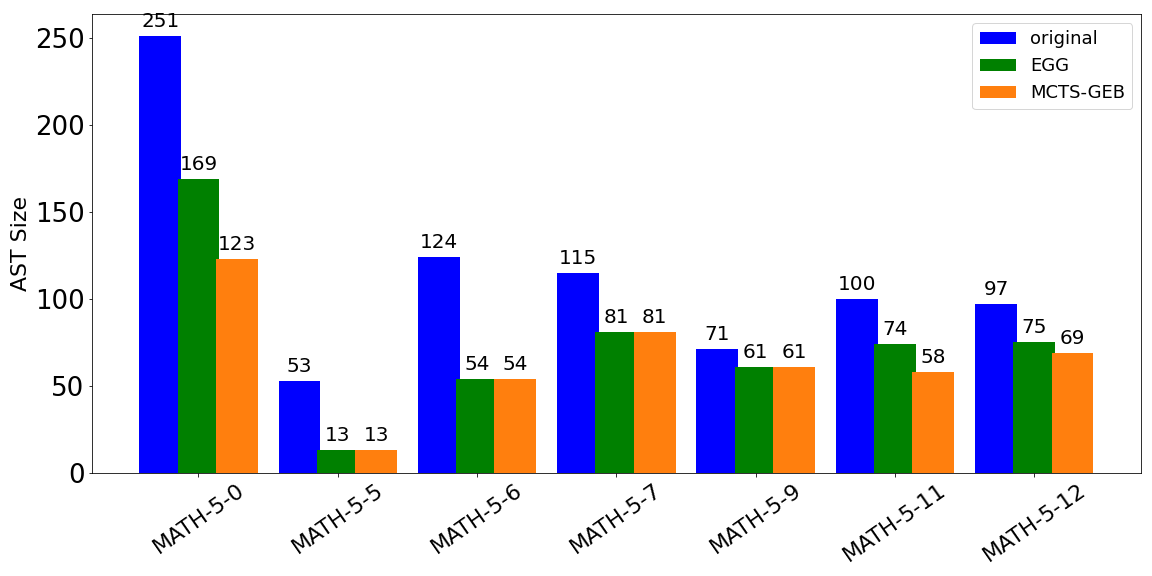}
  \caption{End-to-end expression simplification in the \textit{Math} domain.}
  \label{fig: math-e2e}
  %\vspace{-2mm}
\end{figure}

\begin{figure}[ht]
  \centering
  \includegraphics[width=\linewidth]{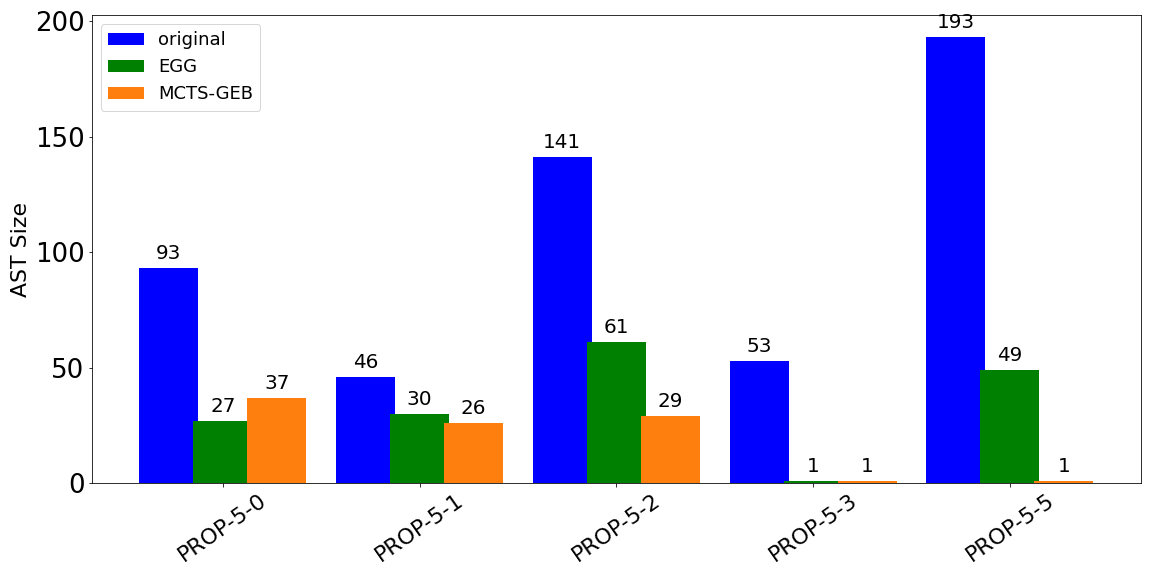}
  \caption{End-to-end expression simplification in the \textit{Prop} domain.}
  \label{fig: prop-e2e}
  %\vspace{-2mm}
\end{figure}

Figure \ref{fig: math-e2e} and \ref{fig: prop-e2e} show the end-to-end comparison of MCTS-GEB vs EGG-based rewrite systems respectively. For \textit{Math} domain, we can observe MCTS-GEB finds shorter expressions in $3$ out of the $7$ initial expressions, while achieving the same performance as the EGG-based system in the rest of the expressions. The biggest improvement is about $1.37$x. In the \textit{Prop} domain, MCTS-GEB outperforms in $3$ out of $5$ expressions, while falling short in $1$ of the expressions. However, the expression reduction is significant, up to $49$x. 

Note that all e-graphs stop because of the node limit, indicating the phase-ordering problem is introduced during the e-graph construction phase. As a result, the experiment results can verify MCTS-GEB is able to plan for better e-graph construction, and therefore, obtain better expressions than EGG-based systems in the two different domains.

\subsection{Optimisation time}

\begin{figure}[ht]
  \centering
  \includegraphics[width=\linewidth]{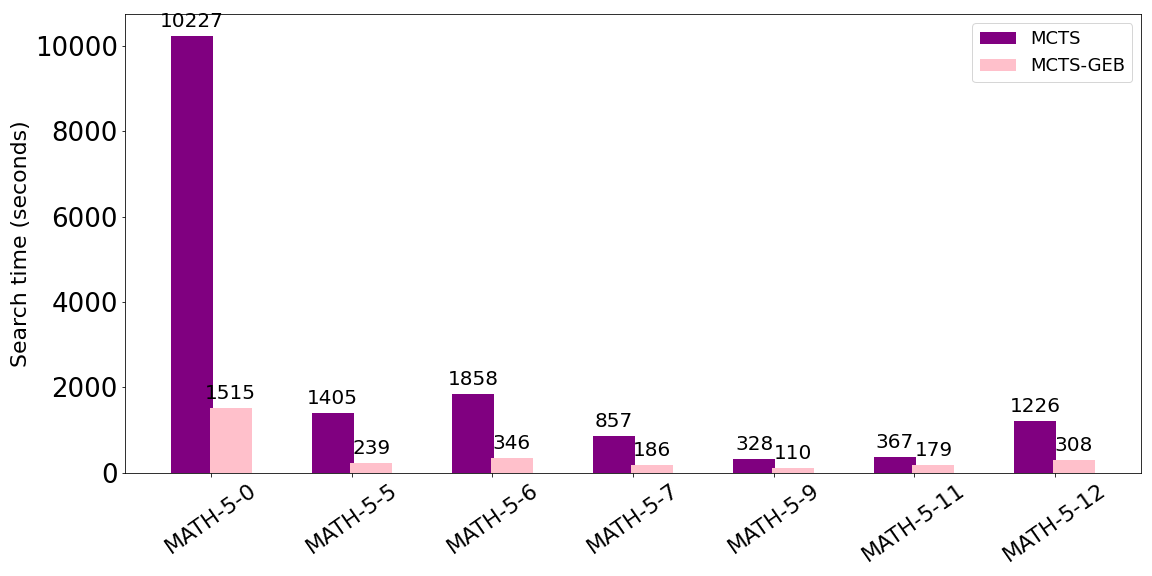}
  \caption{Optimisation time needed in the \textit{Math} domain. The "MCTS" indicates a vanilla MCTS implementation, while "MCTS-GEB" employs a parallel MCTS approach.}
  \label{fig: math-time}
  % \vspace{-5mm}
\end{figure}

\begin{figure}[ht]
  \centering
  \includegraphics[width=\linewidth]{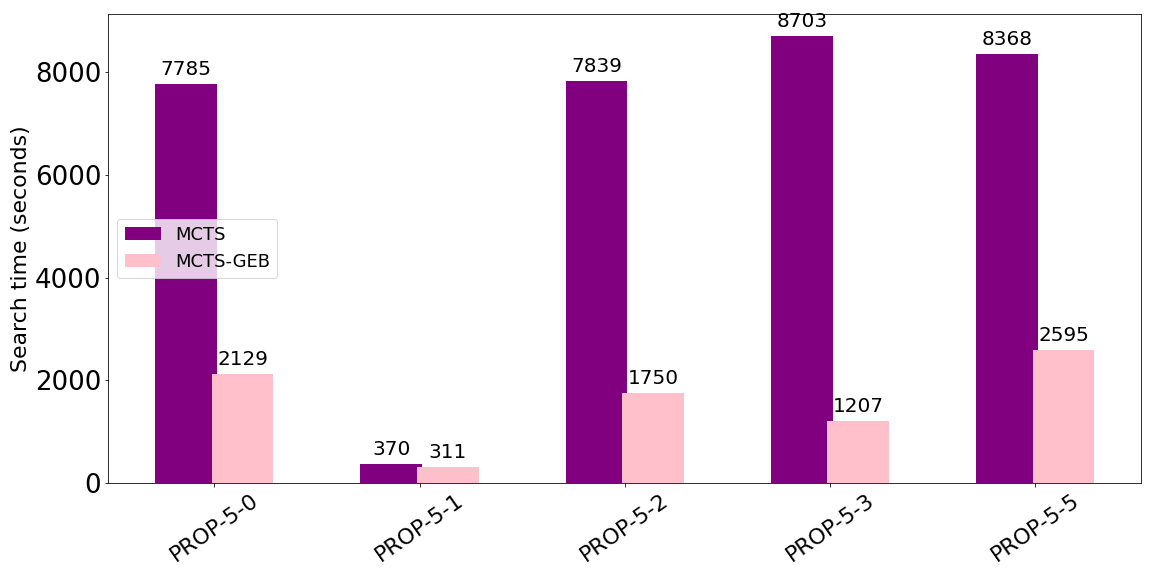}
  \caption{Optimisation time needed in the \textit{Prop} domain. The "MCTS" indicates a vanilla MCTS implementation, while "MCTS-GEB" employs a parallel MCTS approach.}
  %\vspace*{-2mm}
  \label{fig: prop-time}
  
\end{figure}

In this section, we examine the optimisation time taken by MCTS-GEB. We can observe from Figure \ref{fig: math-time} and \ref{fig: prop-time} that using a parallel MCTS algorithm can result in about $6$x less optimisation time in both domains. On average, the optimisation can be finished within $400$ seconds for the \textit{Math} domain and takes about $2000$ seconds for the \textit{Prop} domain. 

However, MCTS-GEB still brings substantial optimisation time overhead compared with EGG-based systems, which are typically finished within seconds, because EGG does not plan for rewrite rules application. We argue that MCTS-GEB introduces a trade-off between better expression extraction and faster optimisation, which is not possible for pure EGG-based systems. Overall, the less-than-an-hour optimisation time is affordable to integrate into systems that are performance-critical and less time-sensitive.

\subsection{Rewrite rule application heatmap}
 
% \begin{figure}
%   \centering
%   \includegraphics[width=\linewidth]{images/egg-math-heatmap.png}
%   \vspace*{-15mm}
%   \caption{Rewrite rule application heatmap of EGG in the \textit{Math} domain. The x-axis denotes rewrite rule indices, while the y-axis denotes different expressions, and the colour bar indicates the application times.}
%   \label{fig: egg-heatmap}
% \end{figure}
% 
% 
% \begin{figure}
%   \centering
%   \includegraphics[width=\linewidth]{images/geb-math-heatmap.png}
%   \vspace*{-15mm}
%   \caption{Rewrite rule application heatmap of MCTS-GEB in the \textit{Math} domain.}
%   \label{fig: geb-heatmap}
% \end{figure}

\begin{figure*}
  \centering
  \includegraphics[width=\linewidth]{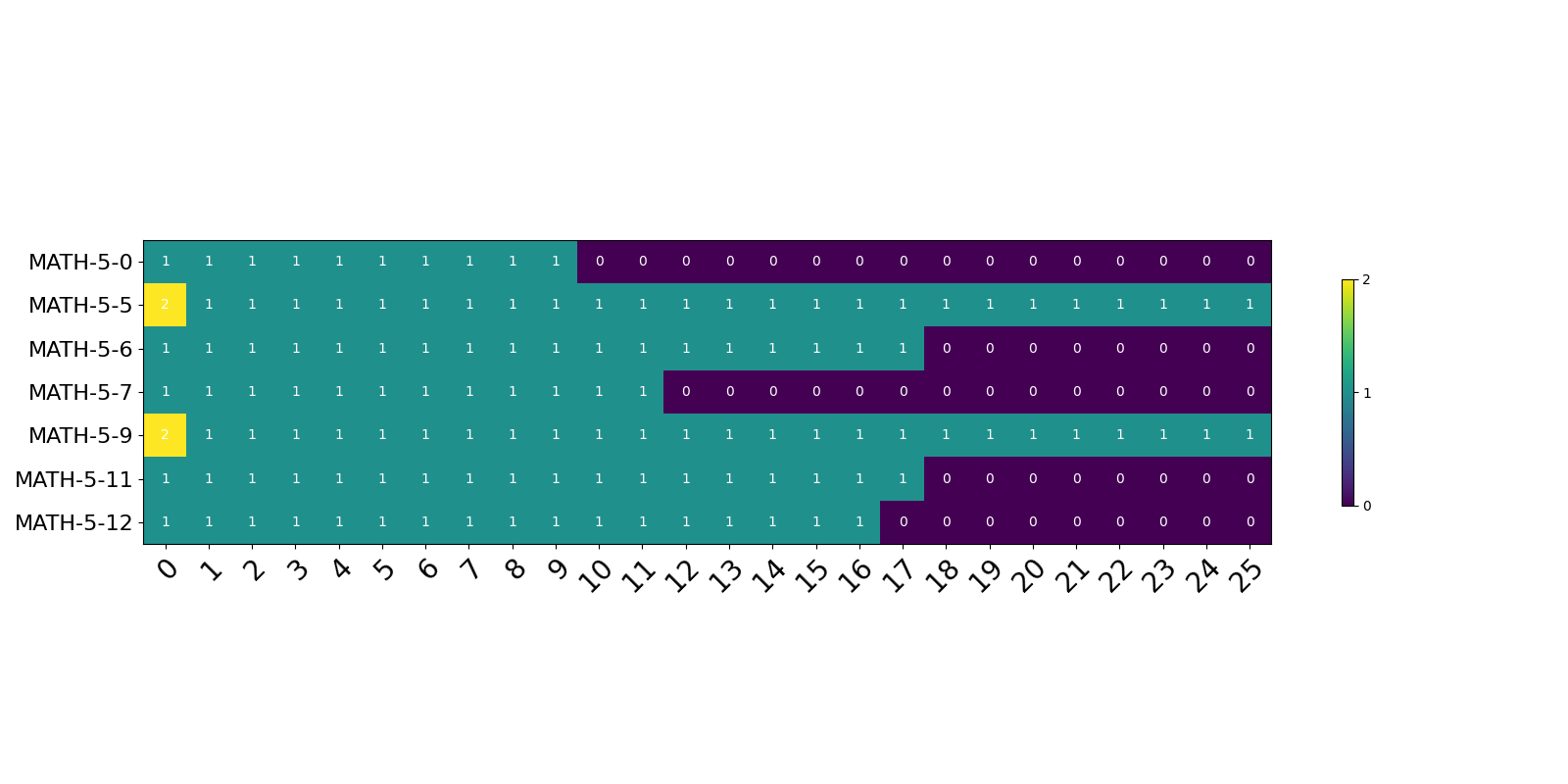}
  \vspace*{-20mm}
  \caption{Rewrite rule application heatmap of EGG in the \textit{Math} domain. The x-axis denotes the rewrite rule indices, while the y-axis denotes different expressions. The number and the associated colour bar indicate the application times of each rewrite rule to every expression.}
  \label{fig: egg-heatmap}
  \vspace*{-5mm}
\end{figure*}

\begin{figure*}
  \centering
  \includegraphics[width=\linewidth]{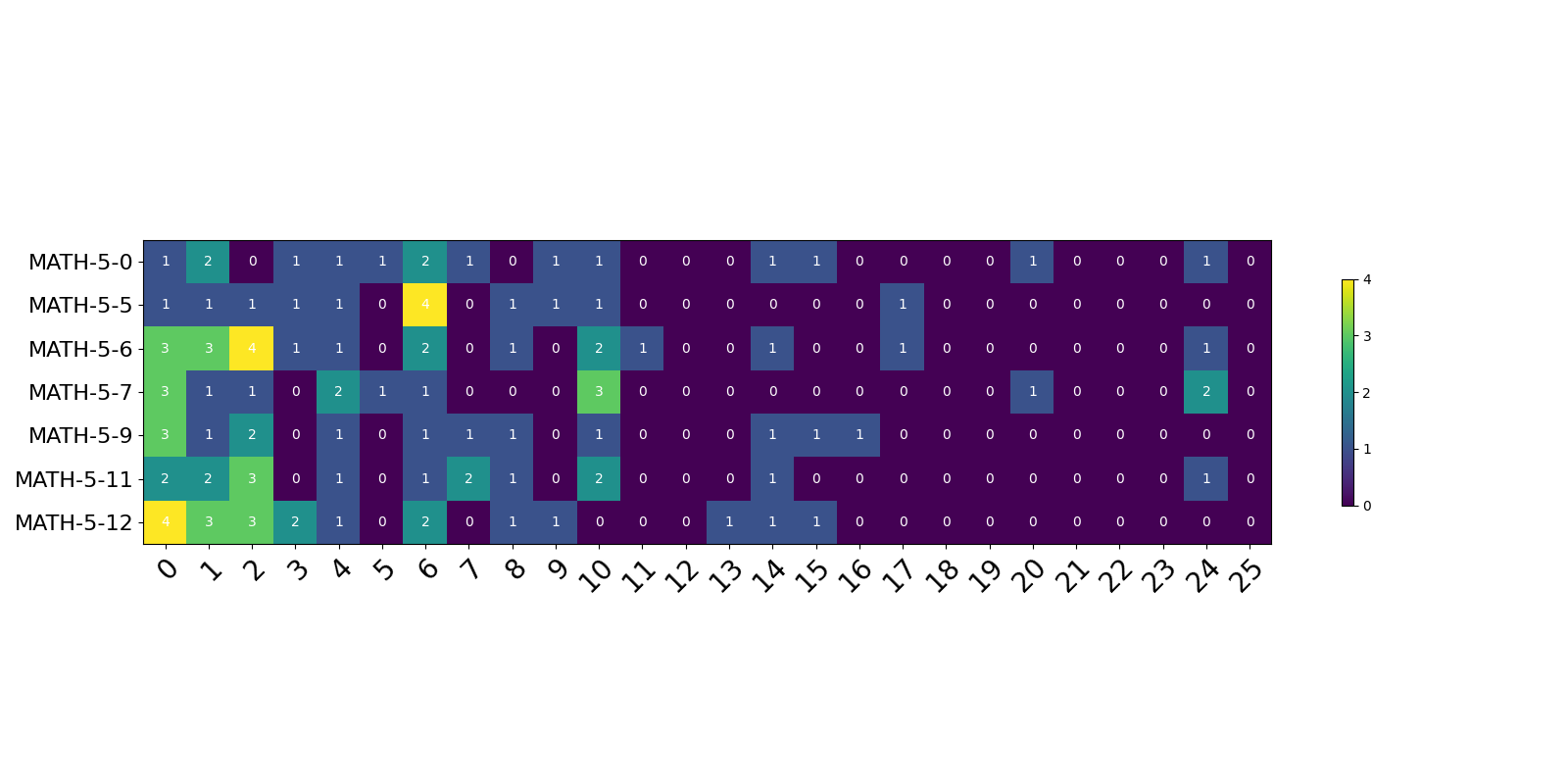}
  \vspace*{-20mm}
  \caption{Rewrite rule application heatmap of MCTS-GEB in the \textit{Math} domain. The axes and numbers have the same meaning as Figure \ref{fig: egg-heatmap}.}
  \label{fig: geb-heatmap}
\end{figure*}

Figure \ref{fig: egg-heatmap} and \ref{fig: geb-heatmap} shows the rewrite rule application heatmaps obtained from running EGG and MCTS-GEB in the \textit{Math} domain respectively. The application times of each rewrite rule to every expression are visualised by the colour bar.

Overall, the heatmap of EGG is more uniform, meaning EGG simply sweeps through available rewrite rules until hitting the node limit during the construction phase. This sometimes leads to sub-optimal performance. For example, when optimising "MATH-5-0", EGG reaches the node limit before it finishes applying all rewrite rules to build the e-graph. 

On contrary, MCTS-GEB can selectively apply rewrite rules, and thus it builds a better e-graph for "MATH-5-0". MCTS-GEB's e-graph is better because the shortest AST has a length of $123$, while it is $169$ for EGG, as shown by Figure \ref{fig: math-e2e}.

\section{Limitation and future work}

The main limitation of MCTS-GEB is its long optimisation time, which is due to the forward planning process and building of the search tree. However, this is inevitable because the phase-ordering problem is NP-hard. As such, while MCTS does not eliminate the phase-ordering problem, it is proven to be effective in previous publications \cite{alphago}.

The optimisation time can be reduced by pruning the action space. For example, if an action sequence does not saturate an e-graph, it indicates the e-graph can represent all possible permutations of the action sequence, and we can prune the action space accordingly. Another possible improvement includes caching the sub-tree of the MCTS search tree at the end of each planning stage, instead of starting from scratch every iteration.

We also utilise multi-core parallelism to accelerate the MCTS process. This indicates we can trade off the computation with the optimisation time to some extent. This means we could add more CPUs to the system and expect a lower optimisation time. For future work, we plan to evaluate the scalability of MCTS-GEB and explore distributed simulation if necessary. For example, we can equip a distributed runtime, such as Ray \cite{ray}, to dispatch simulation tasks over multiple machines and reduce the optimisation time further. 

\section{Conclusion}

We introduce MCTS-GEB, a domain-general rewrite system that uses MCTS for efficient and optimised equality saturation. We explain our motivation in detail and elaborate on the architecture of MCTS-GEB. We conduct experiments in two different domains provided by the EGG library and show MCTS-GEB can outperform the state-of-the-art by $49$x, while optimisation time is generally acceptable. In the future, we plan to further reduce the optimisation time of MCTS-GEB by pruning action space and exploiting distributed computing.  As such, we show the feasibility of MCTS-GEB becoming the building block of the future generation of rewrite systems. 

%%
%% The acknowledgements section is defined using the "acks" environment
%% (and NOT an unnumbered section). This ensures the proper
%% identification of the section in the article metadata, and the
%% consistent spelling of the heading.
%\begin{acks}
%To Robert, for the bagels and explaining CMYK and colour spaces.
%\end{acks}

%%
%% The next two lines define the bibliography style to be used, and
%% the bibliography file.
\bibliographystyle{ACM-Reference-Format}
\bibliography{sample-base}

%%
%% If your work has an appendix, this is the place to put it.
% \appendix
% 
% \section{Research Methods}
% 
% \subsection{Part One}
% 
% Lorem ipsum dolor sit amet, consectetur adipiscing elit. Morbi
% malesuada, quam in pulvinar varius, metus nunc fermentum urna, id
% sollicitudin purus odio sit amet enim. Aliquam ullamcorper eu ipsum
% vel mollis. Curabitur quis dictum nisl. Phasellus vel semper risus, et
% lacinia dolor. Integer ultricies commodo sem nec semper.
% 
% \subsection{Part Two}
% 
% Etiam commodo feugiat nisl pulvinar pellentesque. Etiam auctor sodales
% ligula, non varius nibh pulvinar semper. Suspendisse nec lectus non
% ipsum convallis congue hendrerit vitae sapien. Donec at laoreet
% eros. Vivamus non purus placerat, scelerisque diam eu, cursus
% ante. Etiam aliquam tortor auctor efficitur mattis.
% 
% \section{Online Resources}
% 
% Nam id fermentum dui. Suspendisse sagittis tortor a nulla mollis, in
% pulvinar ex pretium. Sed interdum orci quis metus euismod, et sagittis
% enim maximus. Vestibulum gravida massa ut felis suscipit
% congue. Quisque mattis elit a risus ultrices commodo venenatis eget
% dui. Etiam sagittis eleifend elementum.
% 
% Nam interdum magna at lectus dignissim, ac dignissim lorem
% rhoncus. Maecenas eu arcu ac neque placerat aliquam. Nunc pulvinar
% massa et mattis lacinia.

\end{document}